\documentclass[conference]{IEEEtran}
\usepackage{latexsym}
\usepackage{amsbsy}
\usepackage{amssymb}
\usepackage{amsmath}
\usepackage{verbatim}
\usepackage{graphicx}
\usepackage{algorithmic}
\usepackage{algorithm}
\usepackage{amsfonts}
\usepackage{mathrsfs}
\usepackage{verbatim}
\usepackage{float}
\usepackage{url}
\usepackage{fancybox}
\usepackage{psfrag}
\usepackage{amsmath}
\usepackage{epsfig}
\usepackage{verbatim}
\usepackage{color}
\usepackage{mathabx}
\usepackage{upgreek}
\usepackage{bm}
\usepackage{dsfont}

\renewcommand{\baselinestretch}{1}

\newcommand{\0}{{\mathbf{0}}}
\newcommand{\1}{{\mathbf{1}}}

\newcommand{\vb}[1]{\verb!{#1}!}

\renewcommand{\abstractname}{Summary}

\newcommand{\al}[1]{\begin{align}#1\end{align}}

\newcommand{\be}{\begin{equation}}
\newcommand{\ee}{\end{equation}}
\newcommand{\beqy}{\begin{eqnarray}}
\newcommand{\eeqy}{\end{eqnarray}}
\newcommand{\beqynn}{\begin{eqnarray*}}
\newcommand{\eeqynn}{\end{eqnarray*}}

\newcommand{\ba}{\begin{array}}
\newcommand{\ea}{\end{array}}
\newcommand{\bmx}{\begin{bmatrix}}
\newcommand{\emx}{\end{bmatrix}}
\newcommand{\bsmx}{\left[\begin{smallmatrix}}
\newcommand{\esmx}{\end{smallmatrix}\right]}
\newcommand{\bmxc}[1]{\left[\begin{array}{@{}#1@{}}}
\newcommand{\emxc}{\end{array}\right]}

\newcommand{\bt}[1]{\begin{tabular}{#1}}
\newcommand{\et}{\end{tabular}}

\newcommand{\bc}{\begin{center}}
\newcommand{\ec}{\end{center}}
\newcommand{\ben}{\begin{enumerate}}
\newcommand{\een}{\end{enumerate}}
\newcommand{\bi}{\begin{itemize}}
\newcommand{\ei}{\end{itemize}}

\newcommand{\calN}{\mathcal{N}}
\newcommand{\calR}{\mathcal{R}}
\newcommand{\calX}{\mathcal{X}}
\newcommand{\calI}{\mathcal{I}}

\newcommand{\Rbb}{{\mathbb{R}}}

\newcommand{\Cbb}{{\mathbb{C}}}

\newcommand{\A}{\mathbf{A}}

\newcommand{\C}{\mathbf{C}}

\renewcommand{\H}{\mathbf{H}}
\newcommand{\I}{\mathbf{I}}

\renewcommand{\P}{\mathbf{P}}
\newcommand{\Q}{\mathbf{Q}}

\renewcommand{\a}{\mathbf{a}}
\renewcommand{\b}{\mathbf{b}}
\renewcommand{\c}{\mathbf{c}}

\newcommand{\e}{\mathbf{e}}
\newcommand{\f}{\mathbf{f}}

\newcommand{\h}{\mathbf{h}}

\newcommand{\n}{\mathbf{n}}

\newcommand{\s}{\mathbf{s}}

\renewcommand{\v}{{\mathbf{v}}}

\newcommand{\x}{{\mathbf{x}}}
\newcommand{\y}{{\mathbf{y}}}

\newcommand{\btheta}{\mathbf{\theta}}

\newcommand{\hbx}{{\hat{\x}}}

\IEEEoverridecommandlockouts

\usepackage{cite}
\usepackage{amsmath,amssymb,amsfonts}
\usepackage{algorithmic}
\usepackage{graphicx}
\usepackage{textcomp}
\usepackage{xcolor}
\usepackage{dsfont}
\usepackage{subcaption}
\def\BibTeX{{\rm B\kern-.05em{\sc i\kern-.025em b}\kern-.08em
    T\kern-.1667em\lower.7ex\hbox{E}\kern-.125emX}}
\begin{document}

\title{GD4: Graph-based Discrete Denoising Diffusion for MIMO Detection \\
}

\author{\IEEEauthorblockN{Qincheng Lu}
\IEEEauthorblockA{\textit{School of Computer Science} \\
\textit{McGill University}\\
Montreal, Canada \\
qincheng.lu@mail.mcgill.ca}
\and
\IEEEauthorblockN{Sitao Luan}
\IEEEauthorblockA{\textit{University of Montreal} \\
\textit{Mila - Quebec AI Institute}\\
Montreal, Canada \\
luansito@mila.quebec}
\and
\IEEEauthorblockN{Xiao-Wen Chang}
\IEEEauthorblockA{\textit{School of Computer Science} \\
\textit{McGill University}\\
Montreal, Canada\\
chang@cs.mcgill.ca}
}

\maketitle

\begin{abstract} 

In wireless communications, 
recovering the optimal solution to the multiple-input multiple-output (MIMO) detection problem is NP-hard. 
Obtaining high-quality suboptimal solutions with a favorable performance-complexity trade-off is particularly challenging in under-determined systems with $N_t$ transmit antennas and $N_r < N_t$ receive antennas. 
Recent diffusion-based MIMO detectors have shown promise,
but they require extensive sampling iterations at inference time,
and their performance degrades in under-determined scenarios.
We propose GD4, 
a graph-based discrete denoising diffusion method for MIMO detection.  
Unlike existing diffusion-based detectors that operate in a continuous relaxed space,
GD4 performs denoising directly in the discrete symbol space and enables fast inference with one or a few denoising evaluations. 
Numerical results show that, 
under a similar inference-time compute budget, 
GD4 produces higher-quality suboptimal solutions than existing diffusion-based detectors and some widely used classical baseline including box-constrained Babai point 
and the $K$-best box-constrained randomized Klein-Babai point in both under-determined and overdetermined settings.

\end{abstract}


\begin{IEEEkeywords}
MIMO detection, machine learning, diffusion model
\end{IEEEkeywords}

\section{Problem Definition}
In a MIMO system with $N_t$ transmitter antennas and $N_r$ receiver antennas, 
the transmitted signal $\x_c^*$ and the received signal $\y_c$ satisfy the following linear model
\begin{equation}
\label{eq:sys-model}
\y_c = \H_c \x_c^* + \n_c,
\end{equation}
where $\H_c \in \Cbb^{N_r \times N_t}$ is assumed to be a known channel matrix, 
the elements of unknown integer vector $\x_c^*$ are i.i.d. uniformly distributed over the constellation set $\widetilde{\calX}_k = \{k_1 + k_2j: k_1, k_2 = \pm1, \pm3, \ldots, \pm(2^k-1) \}$,
and the complex circular Gaussian noise $\n_c \sim \mathcal{CN}(\0, \sigma_c^2 \I)$ for some $\sigma_c > 0$. Define
\al{
&\y_r = \begin{bmatrix} \calR(\y_c) \\ \calI(\y_c) \end{bmatrix}, \
\H_r = \begin{bmatrix} \calR(\H_c) & -\calI(\H_c) \\ \calI(\H_c) & \calR(\H_c) \end{bmatrix}, \ 
\x_r^* = \begin{bmatrix} \calR(\x_c^*) \\ \calI(\x_c^*) \end{bmatrix}, \nonumber \\
&\n_r = \begin{bmatrix} \calR(\n_c) \\ \calI(\n_c) \end{bmatrix}, \ 
\bar{\calX}_k = \{ \pm 1, \cdots, \pm (2^k -1) \}, \label{def:ils-real}
}
where $\calR(\cdot)$ and $\calI(\cdot)$ denote the real and imaginary parts. Then the complex system model in~\eqref{eq:sys-model} becomes
\begin{equation} 
\y_r = \H_r \x_r^* + \n_r, \ \ 
\n_r \sim \calN(\0, \sigma_n^2\I), \label{eq:sys-model-real}
\end{equation}
where $\x_r^* \in \calX_k^{2N_t}, \H_r \in \Rbb^{2N_r\times 2N_t}$, $\sigma_n^2=\sigma_c^2/2$. 
The goal is to detect $\x_r$, given $\y_r, \H_r, \sigma_n$.
When $\H_r$ has full column rank, 
the detection problem is said to be overdetermined, 
otherwise it is said to under-determined.

The maximum likelihood detection method
solves the integer least squares (ILS) problem:
\begin{equation}
\min_{\x_r \in \bar{\calX}_k^n} \left\| \y_r - \H_r \x_r \right\|^2_2,
\label{eq:ils}
\end{equation}
which is NP-hard. 
In general, 
an under-determined ILS problem is  much 
more difficult to solve than an overdetermined problem with the
same dimension (i.e., $2N_t$).
An alternative way for the  under-determined problem
is to solve the $L_2$-regularized ILS problem:
\begin{equation}
\min_{\x_r \in \bar{\calX}_k^n} \left \|\bmx \y_r \\ \0 \emx - \bmx \H_r \\ \lambda \I \emx \x_r \right\|^2_2,
\label{eq:ils-l2}
\end{equation}
where $\lambda = \sigma_n / \sigma_x$ with $\sigma^2_x = (2^{k+1} -1)/3$ being the variance of any entry of $\x_r^*$.
This regularized ILS problem is motivated by the linear minimum mean square error (LMMSE) detection.

Finding the optimal solutions to \eqref{eq:ils} and \eqref{eq:ils-l2} are 
often computationally expensive. 
In practice, one often finds their corresponding box-constrained Babai points,
which are referred to as zero-forcing successive interference cancellation (ZF-SIC) detectors and widely used in practice 
\cite{Bab86, WinFH04, BaiCY14, chang2023success}. 
For the unconstrained case, Klein proposed a randomized variant of the Babai point that improves performance by repeating the stochastic process multiple times and choosing the ``best" one~\cite{Kle00}.
The method can readily be extended to the box-constrained case,
leading to the $K$-best box-constrained randomized Klein-Babai point \cite{Wan25},
which will hereafter be referred to as the $K$-best Klein-Babai point for brevity.
There are other methods which find sub-optimal detectors 
such as the FCSD method~\cite{barbero2006performance} 
and the extended Babai method~\cite{chang2024extended}.

In this paper, 
we propose GD4, 
a graph-based discrete denoising diffusion method that enables fast inference for MIMO detection in both under-determined and overdetermined settings. 
Diffusion-based methods generate samples from a target distribution through
a predefined forward process that gradually corrupts the data into noise and a reverse process to recover the original data.
For MIMO detection,
recent works have adopted score-based diffusion methods to sample from the intractable posterior distribution of the transmitted signal given the received signal~\cite{zilberstein2022annealed, he2024massive}.
Here, the score is the gradient of the log-posterior and indicates how to move a candidate solution toward regions of higher posterior probability.
The first diffusion-based MIMO detector employs annealed Langevin dynamics (ALD)~\cite{zilberstein2022annealed} and computes a closed-form posterior score obtained using the SNIPS framework, which relies on the singular value decomposition (SVD) of the channel matrix~\cite{kawar2021snips}.
More recently,
approximate diffusion detection (ADD) is proposed as an SVD-free alternative, 
which approximates the posterior score by combining an iterative detector with Tweedie's identity rather than deriving it analytically~\cite{he2024massive}.
These methods define the forward and reverse process in a continuous relaxed space.
However, they require extensive sequential sampling iterations at inference time, 
which leads to high latency~\cite{luong2025diffusion}.
Moreover, their effectiveness in the under-determined case remains unclear.
On the other hand, 
recent advances in discrete denoising diffusion 
have shown strong performance in discrete generative modeling,
such as text generation and combinatorial optimization~\cite{austin2021structured, sun2023difusco, lou2023discrete}.
To the best of our knowledge, discrete denoising diffusion has not yet been explored for MIMO detection.
The GD4 detector achieves significantly better performance than both classical baselines and diffusion-based methods under similar runtime.

To simplify the presentation of our method, 
we would like to change the constraint set 
$\bar{\calX}_k$ in \eqref{eq:ils} to $\calX_k = \{ 1, \ldots, 2^k\}$.
To do so, we define 
\be \label{eq:ils2}
\begin{split}
& \x  = (\x_r + (2^k + 1) \e)/2, \ \x^*  = (\x_r^* + (2^k + 1) \e)/2,  \\
& \y= \y_r + (2^k + 1) \H_r \e, \ \H = 2 \H_r,  
\end{split}
\ee
where $\e = [1, \dots, 1]^\top$.
Then the linear model \eqref{eq:sys-model-real} becomes 
\be
\y = \H\x^* + \n_r, \label{eq:lm-transformed}
\ee
and the ILS problem \eqref{eq:ils} becomes
\begin{equation}
\min_{\x \in \calX_k} \left\| \y - \H \x \right\|^2_2.
\label{eq:ils-transformed}
\end{equation}

{\bf Notation.} For a matrix $\A$,  denote its $i$-th column by $\a_i$
or $\A_{:,i}$. 
For a vector $\a$, denote its $i$-th element by $\a(i)$ or $a_i$.
Use $[\cdot; \cdot]$ to denote column-wise concatenation, i.e.,
$[\a;\b]=[\a^\top,\b^\top]^\top$.
We write $\x \in \calX_k^{2N_t}$ for a $2N_t$-dimensional vector with entries in $\calX_k$.
Here $\calX_k = \{1, \ldots, 2^k \}$,
we denote by $\1(\x)\in \{0,1\}^{2N_t \times 2^k}$ the one-hot encoding matrix of $\x \in \calX_k^{2N_t}$, 
where $\1(\x)(i, j) = 1$ 
if $\x(i) = j$ and $0$ otherwise. 
For a random vector $\x \in \calX_k^{2N_t}$, 
we define its probability matrix $\P\in \Rbb^{2N_t \times 2^k}$ 
by $\P(i, j)=\Pr(\x(i) = j)$.
We write $\mathrm{Cat}(\x; \P)$ for the categorical distribution on $\calX_k^{2N_t}$ 
with probability matrix $\P$.
We use $\odot$ to denote element-wise multiplication.

\section{Methodology}

\begin{figure*}[t]
\centering
\includegraphics[width=0.8\textwidth]{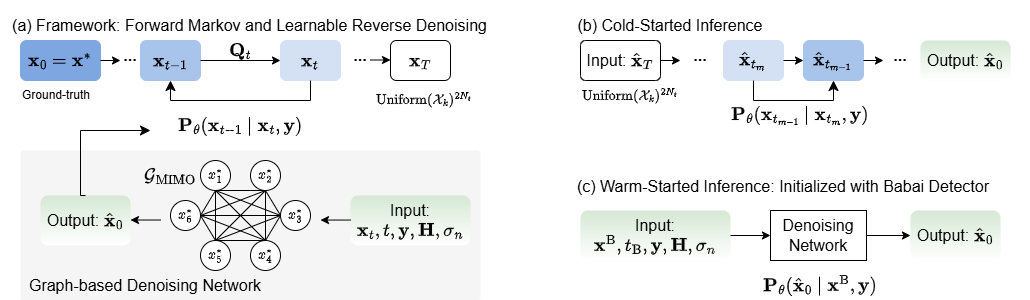}
\caption{Framework of the Proposed Discrete Diffusion-based MIMO Detection. }  
\label{fig:dm_mimo}
\end{figure*}

The diffusion model is characterized by a predefined forward corruption Markov process and a learnable reverse denoising process~\cite{sohl2015deep, ho2020denoising, song2020score}.
For MIMO detection,
the target distribution is the unknown posterior $p(\x^* | \y)$. 
In our formulation,
the forward corruption process is independent of the problem instance,
while the reverse denoising process is parameterized directly as a conditional model given $\y$.

Specifically,
for a training sample with ground-truth transmitted signal $\x_0 = \x^*$, 
the forward process generates a sequence of latent variable $\x_{1:T} = (\x_1, \ldots, \x_T)$ with progressively increasing noise,
so that the terminal distribution approaches $\mathrm{Uniform}(\calX_k)^{2N_t}$, i.e., $\P(\x_T) = \{ 2^{-k}\}^{2N_t \times 2^k}$. 
The denoising network,
parameterized by $\btheta$,
is trained to model the conditional reverse distribution $\P_{\btheta}(\x_{t-1} |\, \x_t, \y)$.
The network parameters are shared across all diffusion steps $t$.
At inference time,
sampling starts from a noisy initialization $\hat{\x}_{T} \sim \mathrm{Uniform}(\calX_k)^{2N_t}$,
which is iteratively denoised by the conditional network to produce an estimate of $\x_0$ for the given $\y$.

We adopt the discrete denoising diffusion model with multinomial noise \cite{austin2021structured, hoogeboom2021argmax},
which in principle allows an arbitrary number of reverse sampling steps at the inference stage. 
On top of this framework,
we propose a Gated Graph Message Passing–based denoising network for MIMO detection,
where the problem instance $(\y,\H, \sigma_n)$ is encoded through graph features.
We further develop two fast inference strategies that require only a few, 
or even a single, 
evaluation of the denoising network to generate high-quality suboptimal solutions.
The following sections describe each component of the framework in Figure~\ref{fig:dm_mimo}.


\subsection{Pre-defined Forward Diffusion Process}
In a discrete diffusion model \cite{ho2020denoising, hoogeboom2021argmax, austin2021structured},
the forward Markov process gradually corrupts the data by adding random noise at each step.
Specifically,
given $\x_{t-1}$, $\x_t$ is generated as:
\begin{equation}
\x_t \sim \mathrm{Cat}\left(\x_t; \1(\x_{t-1}) \Q_t\right),
\label{eq:forward}
\end{equation}
where $\Q_t \in \Rbb^{2^k \times 2^k}$ is the  transition matrix at step $t$.
The matrix $\Q_t$ is applied independently to each entry of $\x_{t-1}$, and $\Q_t(a,b)$ denotes the probability of transitioning from state $a$ at step $t-1$ to state $b$ at step $t$.
Equivalently, $\x_t$ can be generated directly from $\x_0$ as: 
\begin{align}
&\x_t \sim \mathrm{Cat}\left(\x_t; \1(\x_0) \widebar{\Q}_t\right), \nonumber \\s
&\widebar{\Q}_t = \Q_1\cdots\Q_{t-1}\Q_t=\widebar{\Q}_{t-1}\Q_t.
\label{eq:forward-multi}
\end{align}
As the off-diagonal mass of the cumulative transition matrix $\widebar{\Q}_t$ increases,
the noise level in $\x_t$ also increases. 
We choose $\Q_t$ so that the noise level introduced by $\widebar{\Q}_t$ increases gradually with $t$,
and the terminal distribution satisfies $\x_T \sim \mathrm{Uniform}(\calX_k)^{2 N_t}$. 
In this way,
$\x_t$ becomes increasingly different from $\x_0$ as
the diffusion step $t$ increases,
while the final $\x_T$ follows the uniform distribution from which reverse sampling can be initialized.
In practice, we construct $\Q_t$ using discretized Gaussian probabilities, 
taking into account the ordinal nature of $\calX_k$. 
The detailed configuration of $\Q_t$ are provided in Section~\ref{sec:exp}. 


\subsection{Reverse Denoising Network}
\label{sec:reverse}
The goal of diffusion-based MIMO detection is to denoise a noisy candidate solution.
Denoising depends on the conditional reverse distribution $\P(\x_{t-1} \,|\, \x_t, \y)$,
which does not have a closed-form expression.
This section describes how $\P(\x_{t-1} \,|\, \x_t, \y)$ is parameterized and approximated using a neural network.

Following \cite{ho2020denoising, hoogeboom2021argmax, austin2021structured},
we use a denoising network parameterized by $\btheta$ to model the conditional distribution $\P_{\btheta}(\hat{\x}_0 | \x_t, \y) \in \Rbb^{2N_t\times 2^k}$,
where the $i$-th row gives the predicted categorical distribution of $\x_0(i)$ over $\calX_k$.
Given this prediction, 
the reverse distribution at step $t$ is given by:
\begin{align}
&\P_{\btheta}(\x_{t-1} \,|\, \x_t, \y) \, \propto \, \P(\x_{t-1} \,|\, \x_t, \hat{\x}_0) \odot
\P_{\btheta}(\hat{\x}_0 \,|\, \x_t, \y) \nonumber \\
&\propto \, \P(\x_t \,|\, \x_{t-1}) \odot \P(\x_{t-1} \,|\, \hat{\x}_0) \odot \P_{\btheta}(\hat{\x}_0 \,|\, \x_t, \y).
\label{eq:reverse}
\end{align}
Based on~\eqref{eq:reverse}, it can be shown that the $i$-th element $\x_{t-1}(i)$ follows the distribution~\cite{austin2021structured}:  
\begin{equation}
\x_{t-1}(i) \sim \mathrm{Cat} 
\left( \x_{t-1}(i); 
\frac{\left( \1(\x_t)_{i, :} \Q_t^\top \right) \odot 
\left( \P_{\btheta}(i, :) \widebar{\Q}_{t-1} \right) }
{\P_{\btheta}(i, :)  \widebar{\Q}_{t} \1(\x_t)_{i, :}^\top} \right),
\label{eq:reverse_element}
\end{equation} 
where $\P_{\btheta}$ is shorthand for $\P_{\btheta}(\hat{\x}_0 | \x_t, \y)$.
Once the prediction $\P_{\btheta}$ is available, 
it defines the reverse posterior used to sample $\x_{t-1}$ from $\x_t$.
Repeating this denoising process from $\x_T$ yields the final estimate of $\x_0$.


\subsection{Gated Graph Message Passing-based Denoising Network}
\label{sec:dm_model}
In this section,
we propose a Gated Graph Message Passing-based denoising network to model $\P_{\btheta}(\hat{\x}_0 \,|\, \x_t, \y)$,
which is used to construct the reverse posterior in~\eqref{eq:reverse}.

While the forward diffusion process corrupts each entry $\x_t(i)$ independently,
the true posterior couples all symbols through the received signal $\y$ and the channel matrix $\H$.
To capture these dependencies,
we represent each problem instance as a fully connected graph,
where nodes correspond to the entries of $\x^*$,
and node and edge features are defined using the instance information $(\y, \H, \sigma)$.
The network then denoises $\x_t$ by performing message passing on this graph.
This graph-based formulation also allows the denoising process to effectively exploit the conditional information.


For a given instance $(\y, \H)$,
we define a fully connected graph with self-loops, 
$\mathcal{G}_{\mathrm{MIMO}} = (\mathcal{V}, \mathcal{E})$,
where each node $i \in \mathcal{V}$ represents the transmitted symbol $x_i^*$.
In $\mathcal{G}_{\mathrm{MIMO}}$, 
an edge $e_{ij}$ exists between each pair of nodes $(i, j)$.
To inject problem-specific inductive bias into the denoising network,
we initialize the node feature $\v_i$, 
and edge feature $\e_{ij}$ according to the pair-wise Markov random field factorization of the posterior~\cite{scotti2020graph}:
\begin{align}
&\v_i = [\y^\top \h_i, \h_i^\top \h_i, \sigma_n^2]^\top, \quad \e_{ij} = [-\h_i^\top \h_j, \sigma_n^2]^\top.
\label{eq:init-feature}
\end{align}
Here, 
the node features encode local evidence associated with symbol $x_i^*$, 
while the edge features encode pairwise interactions induced by the channel matrix.

In addition to the instance graph, 
the denoising network must also incorporate the current diffusion state.
To this end, we encode the noisy sample $\x_t$ and the denoising step $t$ as graph signals.
We first map elements of $\x_t$ to a continuous range and add a small multiplicative perturbation:
\begin{equation}
\s(i) = \left( \frac{2\x_t(i)-2}{2^k -1} - 1 \right)(1 + 0.05 {\epsilon_i}), \ \
\epsilon_i \sim \mathcal{N}(0, 1),
\label{eq:normalize}
\end{equation}
for $i=1:2N_t$,
so that they lie approximately in $[-1, 1]$. 
We then apply sinusoidal embeddings~\cite{vaswani2017attention} to both $\s(i)$ and $t$.
Specifically, define the sinusoidal map $\f(\cdot; \tau, H): \Rbb \rightarrow \Rbb^{H}$ by:
\begin{align}
&\f(x; \tau, H)_{2j-1} = \mathrm{sin} \left( x / (\tau)^{2j/H} \right), \nonumber \\
&\f(x; \tau, H)_{2j} = \mathrm{cos} \left( x / (\tau)^{2j/H} \right),
\end{align}
where $\tau$ is the temperature, $H$ is the size of hidden dimension, $j=1:\frac{H}{2}$.
The noisy symbol embedding 
$\C \in \Rbb^{H \times 2N_t}$ and step embedding
$\mathbf{t} \in \Rbb^{H}$ are defined as:
\begin{equation}
\c_i = \f(\s(i); 10000, H), \quad \mathbf{t} = \f(t; 10000, H),
\end{equation}
where $\c_i$ is the $i$-th column of $\C$.
These embeddings provide smooth representations of the noisy discrete symbols and the diffusion step, 
allowing the network to generalize across denoising stages.

We next describe the gated graph message passing at each layer of the denoising network.
For each node $i$ and edge $(i, j)$, 
first compute the initial node embedding $\v_i^{(0)}$ and the edge embedding $\e_{ij}^{(0)}$ as:
\begin{align}
&\v_i^{(0)} = \mathrm{MLP}_1\left( [\mathrm{MLP}_2(\v_i); \c_i] \right) \in \Rbb^{H}, \label{eq:xt_feature}\\
&\e_{ij}^{(0)} = \mathrm{MLP}_{\mathrm{e}}\left( \e_{ij} \right) \in \Rbb^{H},
\end{align}
where each $\mathrm{MLP}$ denotes a linear transformation.
At layer $l$,
a relational embeddings $\mathbf{r}_{ij}^{(l)}$ is computed for each node pair $(i, j)$:
\begin{align}
\mathbf{r}_{ij}^{(l)} = \mathrm{MLP}_r^{(l)}\left([\e_{ij}^{(l)}; \v_i^{(l)}; \v_j^{(l)}] \right).
\end{align}
This relational embeddings determine how strongly node $j$ should influence node $i$,
and it allows the aggregation weights to depend on both endpoints and their interaction.
The gated neighborhood aggregation is defined as:
\begin{equation}
\hat{\v}_i^{(l)} = \mathrm{MLP}_3^{(l)} (\v_i^{(l)}) + \sum_j \sigma \left( \mathbf{r}_{ij}^{(l)} \right) \odot \mathrm{MLP}_4^{(l)} (\v_i^{(l)}),
\end{equation}
where $\sigma(\cdot)$ denotes the element-wise sigmoid function.

Finally,
the edge and node embeddings are updated using residual connections~\cite{he2016deep} and denoising steps conditioning:
\begin{align}
&\v_i^{(l+1)} = \v_i^{(l)} + \mathrm{ReLU} \left( \hat{\v}_i^{(l)} \right) + \mathrm{MLP}_{t}^{(l)}( \mathbf{t} ), \label{eq:t_feature} \\
&\e_{ij}^{(l+1)} = \e_{ij}^{(l)} + \mathbf{r}_{ij}^{(l)}.
\end{align}
The term $\mathrm{MLP}_t^{(l)}(\mathbf{t})$ injects the denoising-step information into each layer. 
This graph message passing on $\mathcal{G}_{\mathrm{MIMO}}$ enables the network to jointly exploit the problem instance $(\y, \H, \sigma)$ and the current diffusion state $(\x_t,t)$.

At the output layer, 
the node embedding $\v_i^{(L)}$ parameterizes the predicted distribution $\P_{\btheta}(\hat{\x}_0 \,|\, \x_t, \y)_{i, :} = [p_1, \ldots, p_{2^k}]^\top$ over $\calX_k$. 
We now describe the process to obtain $\{ p_c \}_{c=1}^{2^k}$.
Since the values in $\calX_k$ are ordinal,
we incorporate this structure by modeling the output distribution with a discretized 
truncated logistic distribution~\cite{austin2021structured}.
Specifically, 
for each $\hat{\x}_0(i)$, 
the network predicts the location $\mu_i$ and the scale $\sigma_i$ of a logistic distribution via:
\begin{align}
&(\hat{\mu}_i, \: \sigma_i) = \mathrm{CNN}\left(\mathrm{ReLU}\left(\v_i^{(L)} \right) \right),\\
&\mu_i = \mathrm{tanh}\left(\hat{\mu}_i + \s(i) \right), \label{eq:loc}
\end{align}
where $\mathrm{CNN}$ is a convolutional layer with two output channels.
In~\eqref{eq:loc}, 
the model learns a residual correction around the 
current noisy $\s(i)$,
while $\mathrm{tanh}(\cdot)$ activation keeps $\mu_i$ within $[-1, 1]$.
Given the predicted location $\mu_i$ and the scale $\sigma_i$, 
we discretize the logistic distribution 
using $2^k$ evenly spaced bin centers:
\begin{equation}
b_c = -1 + \frac{2 (c-1)}{2^k - 1},\quad c = 1, \ldots, 2^k,
\end{equation}
with bin width $\Delta = \frac{2}{2^k - 1}$.
Accordingly, the bin boundaries for class $c$ are
\begin{equation}
b_c^{-} = b_c - \frac{\Delta}{2}, \quad b_c^{+} = b_c + \frac{\Delta}{2}.
\end{equation}
The probability for class $c$ is then computed as $p_c = F(b_c^{+}; \mu_i, \sigma_i) - F(b_c^{-}; \mu_i, \sigma_i)$, where $F$ denotes the CDF of the logistic distribution.
The resulting probabilities $\{ p_c \}_{c=1}^{2^k}$ concentrate on classes near $\mu_i$ and decay smoothly as the classes move farther away.
This design assigns higher probability to ``near-miss" predictions than to distant ones,
which is consistent with the ordinal structure of $\calX_k$ and leads to improved model performance.


\begin{algorithm}[t]
\caption{Training of the GD4 denoising network $f_{\btheta}$}
\label{alg:gd4_training_loss}
\begin{algorithmic}[1]
\FOR{each training iteration}
\STATE Generate a problem instance $(\H,\y,\sigma, \x_0=\x^*)$
\STATE Sample $t \sim \mathrm{Uniform}(\{1,\ldots,T\})$
\STATE Sample $\x_t \sim \P(\x_t \,|\, \x_0)$ via~\eqref{eq:forward-multi}
\STATE Compute $\P_{\btheta}(\hat{\x}_0 \,|\, \x_t,\y) = f_{\btheta}(\x_t, \y, \H, \sigma, t)$
\STATE Compute $\P_{\btheta}(\x_{t-1} \,|\, \x_t,\y)$ via~\eqref{eq:reverse}
\STATE Take gradient descent step on $\nabla_{\btheta} \mathcal{L}$ where
\begin{align*}
&\mathcal{L}_{\mathrm{vb}}
=
\sum_i
\mathrm{KLD}\!\left(
\P(\x_{t-1} |  \x_t,\x_0)_{i,:}
\,\middle\|\,
\P_{\btheta}(\x_{t-1} | \x_t,\y)_{i,:}
\right)\\
&\mathcal{L}_{\mathrm{ce}}
=
-\mathrm{trace}\!\left(
\1(\x_0)^\top
\log \P_{\btheta}(\hat{\x}_0 \,|\, \x_t,\y)
\right)\\
&\mathcal{L} = \mathcal{L}_{\mathrm{vb}} + \mathcal{L}_{\mathrm{ce}}
\end{align*}
\ENDFOR
\end{algorithmic}
\end{algorithm}

\subsection{Optimization}
Let $f_{\btheta}(\x_t, \y, \H, \sigma)$ denote the denoising network proposed in Section~\ref{sec:dm_model}.
Algorithm~\ref{alg:gd4_training_loss} describes how its parameters $\btheta$ are optimized.
In Algorithm~\ref{alg:gd4_training_loss}, 
the variational loss $\mathcal{L}_{\mathrm{vb}}$ minimizes the Kullback--Leibler divergence between the true reverse posterior induced by the forward process and the model-predicted reverse distribution.
Note that the true reverse posterior $\P(\x_{t-1}\,|\, \x_t,\x_0)_{i, :}$ can be computed in closed form by replacing the predicted term $\P_{\theta}$ in ~\eqref{eq:reverse_element} with $\1(\x_0)$.
The cross-entropy loss $\mathcal{L}_{\mathrm{ce}}$ encourages accurate prediction of the clean signal $\x_0$ from the noisy state $\x_t$~\cite{austin2021structured, sun2023difusco}.

\subsection{Cold-Start Inference: Step-skipped Reverse Sampling Initialized with $\x_T \sim \mathrm{Uniform}(\calX_k)^{2N_t}$}
\label{sec:infer1}
To accelerate inference, 
we reduce the number of reverse denoising steps.
We refer to this strategy as cold-started detection. 
Instead of traversing the full reverse chain,
cold-started detection uses $M << T$ reverse steps along a shortened reverse chain $\x_{t_M}, \ldots, \x_{t_1}$, $t_M = T$ and $t_1 = 0$~\cite{song2021denoising}. 
The procedure starts from $\x_T \sim \mathrm{Uniform}(\calX_k)^{2N_t}$,
and employs step-skipping to reduce the number of denoising network evaluations (Fig.~\ref{fig:dm_mimo} (b)).
To support step-skipping,
we consider the reverse distribution $\P_{\btheta}(\x_{t_1} \,|\, \x_{t_2}, \y)$ for any $t_1 < t_2$. 
Its form can be derived analogously to~\eqref{eq:reverse}, 
except that $\widebar{\Q}_{t-1}$ is replaced by the cumulative transition matrix $\widebar{\Q}_{t_1, t_2} = \Q_{t_1 + 1}, \ldots \Q_{t_2}$ instead of $\widebar{\Q}_{t}$,
which is precomputed and shared across all test sample.

\subsection{Warm-Start Inference: One-Step Denoising Initialized with the Babai Detector}
\label{sec:infer2}

The denoising process does not have to start from a non-informative sample $\x_T$.
Instead,
it can be warm-started from a sub-optimal solution.
Specifically,
our numerical tests show that the trained denoising network can effectively refine the Babai point and produce a higher-quality estimate with a single denoising network evaluation, as illustrated in Fig.~\ref{fig:dm_mimo}(c).

For the ILS problem, 
we estimate in advance the average symbol error rate (SER) of the Babai point at each signal-to-noise ratio (SNR).
The pre-defined forward Markov induces an expected SER for $\x_t$ at each diffusion step $t$.
Therefore, given the SNR,
we select $t_{\mathrm{\scriptscriptstyle B}}$ such that the expected SER of $\x_{t_{\mathrm{\scriptscriptstyle B}}}$ approximately matches the expected SER of the Babai point.
Let $\hbx_r^{\mathrm{\scriptscriptstyle B}}$ denote the Babai point corresponding 
to \eqref{eq:ils} for the overdetermined case or \eqref{eq:ils-l2} for 
the under-detremined case. 
We then transform it to $\hbx^{\mathrm{\scriptscriptstyle B}}$ according to \eqref{eq:ils2}. 
Warm-started inference performs one-step denoising via
$\P_{\btheta}(\hat{\x}_0 \,|\, \hat{\x}^{\mathrm{\scriptscriptstyle B}}, \y)$.
We use this one-step prediction as the final detection result. 

\section{Numerical Tests}
\label{sec:exp}

\subsection{Experimental Setup}

This section compares the MIMO detection performance of the proposed the GD4 detector with 
existing diffusion-based method including ALD~\cite{zilberstein2022annealed} and ADD~\cite{he2024massive},
and classical baselines including the box-constrained Babai detector~\cite{WinFH04}
and the K-best box-constrained randomized Klain-Babai detector~\cite{Wan25}.

For each problem instance,
the entries of $\H_c$ are drawn i.i.d. from $\mathcal{CN}(0, 1/N_{r})$.
The $\x_{r}^{*}$ is sampled uniformly from $\mathcal{X}_k^{2N_t}$ with $\sigma_x^2 = (2^{k+1} - 1)/3$.
The $\mathrm{SNR}_{\mathrm{db}}$ is:
\begin{equation}
\mathrm{SNR}_{\mathrm{db}} = 10 \mathrm{log}_{10} \left( \mathbb{E} [\|\H_c \x_c \|_2^2 ] / \mathbb{E}[\| \n_c \|_2^2] \right).
\label{eq:snr}
\end{equation}
Given a target $\mathrm{SNR}_{\mathrm{db}}$,
we use the corresponding $\sigma_n$ to generate $\n_r$ and compute $\y_r$.
We train the denoising network using
$\x^{*}$, $\y$, $\H$ in~\eqref{eq:ils2}. 
For a given $\x_0 = \x^{*}$ and diffusion step $t$, 
we adopt the discretized Gaussian transition matrix $\Q_t$~\cite{austin2021structured} to compute $\x_t$ in~\eqref{eq:forward-multi}:
\begin{equation}
\Q_t(i,j) = \begin{cases} 
\frac{\exp\left(- 4|i-j|^2 / (2^k-1)^2 \beta_t \right)}{\sum_{m=1}^{2^k} \exp \left( -4m^2 / (2^k-1)^2 \beta_t \right)}, & i\neq j, \\
1 - \sum_{m \neq i} \Q_t(i,m), & i=j.
\end{cases}
\end{equation}
Here, a linear noise schedule $\{ \beta_t \}_{t=1}^T = \mathrm{linspace}(\beta_1, \beta_T, T)$ with $\beta_1 = 8.5\times 10^{-3}$, $\beta_T = 1.7\times10^{-2}$, $T=1000$ is used.
The network
is trained by $380,000$ iterations to minimize the training loss.
Each iteration uses 32 randomly generated instances,
with $\mathrm{SNR}_{\mathrm{dB}}$ sampled uniformly from $[30, 40]$.
The training use the Adam optimizer with a learning rate $10^{-4}$ and weight decay $5 \times 10^{-5}$.
We use hidden size $H = 32$ and $L=12$ layers for all experiments.
We do not observe overfitting and therefore use the trained network at the last iteration for evaluation.

For testing, all methods are evaluated on $10^5$ randomly generated instances for each $\mathrm{SNR}_{\mathrm{dB}}$.
In the under-determined case,
Babai and K-best randomized Babai are applied to the regularized ILS problem in~\eqref{eq:ils-l2},
which is required.
The proposed GD4, ALD and ADD works directly on~\eqref{eq:ils2} without adding regularization for all cases.

\subsection{Complexity and Runtime Analysis}
Table~\ref{tab:runtime} reports the average per-instance detection runtime.
All methods are implemented with GPU acceleration,
and the $K$ Klein-Babai candidates are produced in parallel.
In all experiments, 
we use a batch size of 32, 
and the samples within each batch are computed in parallel.
Let $m = 2 N_r$, $n = 2 N_t$.
Computing the Babai point has complexity $\mathcal{O}(mn^2)$, 
while computing the $K$-best Klein-Babai 
is of $\mathcal{O}(mn^2 + K n^2)$. 
The complexity of ADD and ALD is dominated by $t$ sequential steps, 
resulting in complexity $\mathcal{O} (t n^2)$.
For a fair runtime comparison,
we use $t = 100$ for ALD and $t = 1000$ steps for ADD.
The complexity of our GD4 method includes $\mathcal{O}(mn^2)$ for graph feature initialization, 
and $\mathcal{O}(HL n^2)$ per denoising network evaluation, where $H$ is the hidden size, and $L$ is the number of layers.
Warm-started detection first computes the Babai point and then performs a single evaluation of the denoising network.
For cold-started detection with $t$ sampling steps, 
the total complexity is $\mathcal{O}(mn^2 + t H L n^2)$.
We test cold-started detection with number of sampling step $t = 1, 3, 10$. 

\begin{table}[htbp]
\caption{Average Detection Runtime ($N_t = 32$, $N_r = 28$)}
\vspace{-3 mm}
\begin{center}
\begin{tabular}{l c}
\hline
Method & Runtime (seconds) \\ \hline
Babai & $6.07 \times 10^{-4}$ \\ 
10-best Klein-Babai & $1.87 \times 10^{-3}$ \\ 
ALD (100 steps) & $2.03 \times 10^{-2}$ \\
ADD (1000 steps) & $1.24 \times 10^{-2}$ \\
GD4: Warm-started (1 step) & $1.12 \times 10^{-3}$\\ 
GD4: Cold-started (1 step) & $4.59 \times 10^{-4}$ \\
GD4: Cold-started (3 steps) & $1.53 \times 10^{-3}$ \\
GD4: Cold-started (10 steps) & $5.12 \times 10^{-3}$\\ \hline
\end{tabular}
\label{tab:runtime}
\end{center}
\end{table}
\vspace{-5mm}

\begin{figure}
  \centering
  \begin{subfigure}{\columnwidth}
  \centering
  \caption{$N_t = 32$, $N_r = 28$}
\includegraphics[width=\columnwidth, height=0.3\textheight]{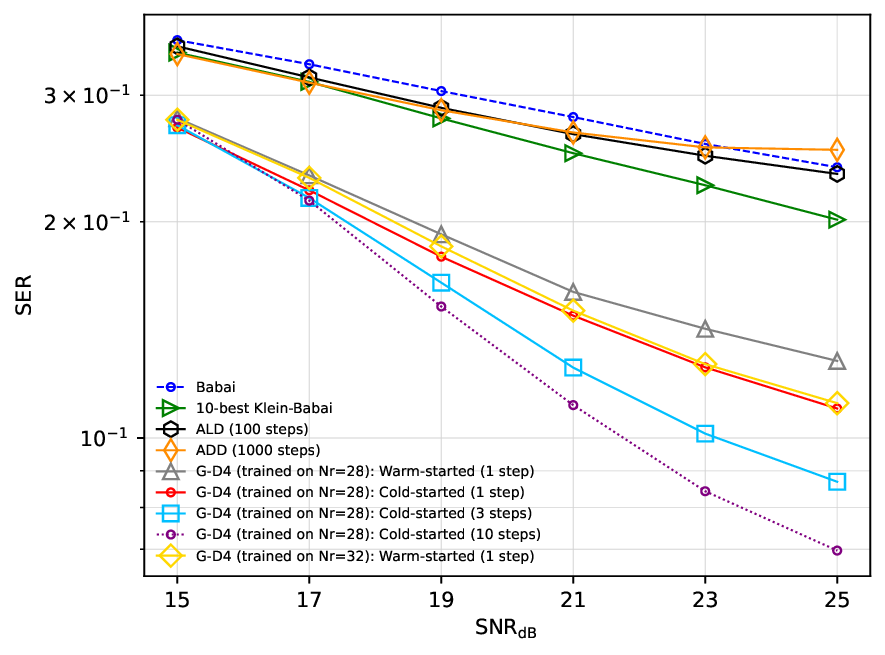}
  \end{subfigure}
  \vspace{-3mm}
  \begin{subfigure}{\columnwidth}
  \caption{$N_t = 32$, $N_r = 30$} 
  \includegraphics[width=\columnwidth, height=0.3\textheight]{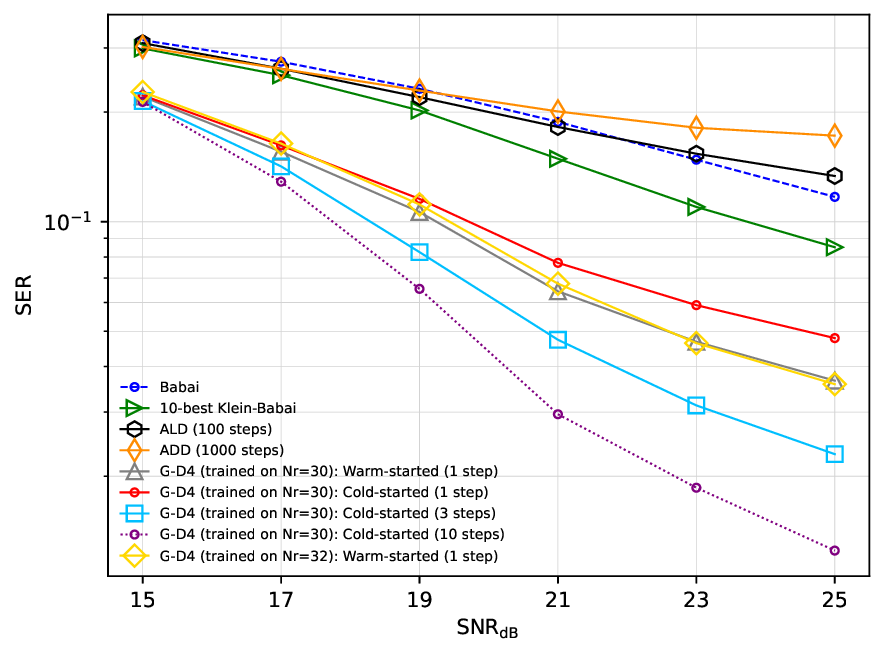}
  \end{subfigure}
  \vspace{-3mm}
  \begin{subfigure}{\columnwidth}
  \caption{$N_t = 32$, $N_r = 32$}
\includegraphics[width=\columnwidth, height=0.3\textheight]{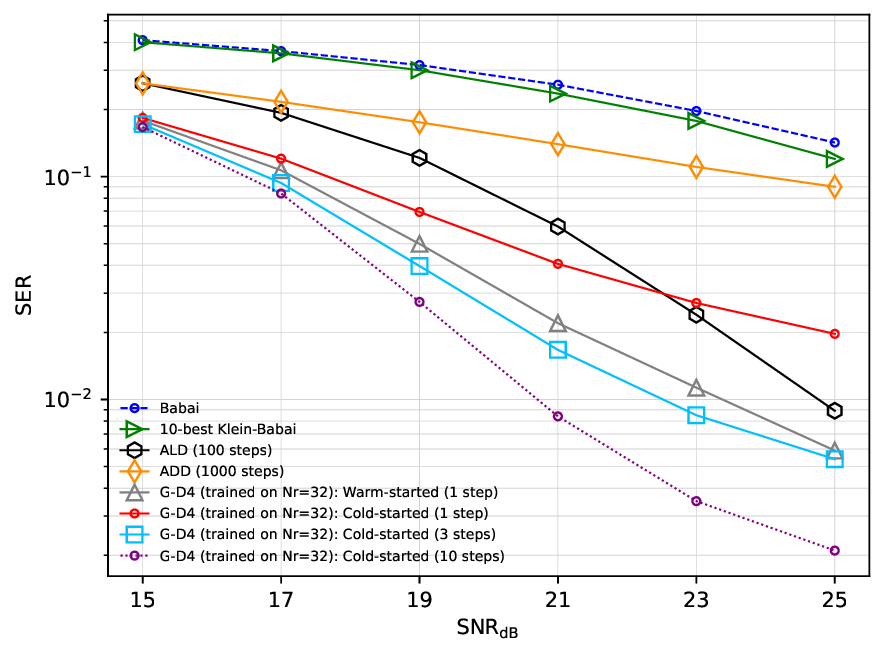}
  \end{subfigure} 
  \caption{Performance Comparison on 16 QAM}
  \label{fig:result}
\end{figure}

\subsection{Performance Comparison}

In Figure~\ref{fig:result}, 
we compare the detection performance of the proposed GD4 with baseline methods for 16-QAM.
Both an overdetermined setting ($N_t = N_r = 32$) 
and under-determined settings ($N_t = 32$, $N_r \in \{ 30, 28\}$) are used.
For GD4,
we report results for warm-started inference, 
initialized from the Babai point using a single denoising step, 
as well as cold-started inference initialized from a uniform noise using 1, 3, and 10 denoising steps.
The runtime of methods shown in Figure~\ref{fig:result} can be found in Table~\ref{tab:runtime}.

We make the following key observations:
(1) With a single denoising step in the warm-started setting, 
GD4 outperforms the 
10-best Klein-Babai point as well as the diffusion-based baselines ALD and ADD, 
while requiring less runtime.
(2) In the under-determined case,
both ALD and ADD become less effective, 
while GD4 remains robust.
(3) The denoising network generalizes well 
across problem sizes.
In particular, 
a network trained on instances with $N_t = N_r = 32$ also performs well on under-determined problems with $N_r = 30$ and $N_r = 28$.
(4) Although a single denoising step is already effective, increasing the number of denoising steps under cold-started inference further improves detection accuracy.

\section{Conclusion}
We introduced GD4, 
a graph-based discrete denoising diffusion method for MIMO detection. 
Unlike existing diffusion-based methods, 
which require many sequential sampling iterations at inference time and degrade substantially in under-determined scenarios, 
GD4 enables fast and effective detection by operating directly in the discrete space. 
We developed two efficient inference strategies,
a cold-started variant that denoises a uniformly sampled initialization, 
and a warm-started variant that refines the Babai point. 
Overall, 
our results demonstrate the effectiveness of discrete denoising diffusion for MIMO detection and motivate future work on extending this framework to broader tasks in communication.

\newpage
\bibliographystyle{IEEEtran}
\bibliography{ref} 

@article{Bab86,
  title={ On {Lov\'{a}sz} lattice reduction and the nearest lattice point problem},
  author={Babai, L.},
  journal={Combinatorica},
  volume={6},
  number={1},
  pages={1-13},
  year={1986},
  publisher={IEEE}
}

@article{WinFH04,
  title={Lattice-Reduction-Aided Broadcast Precoding},
  author={Windpassinger, C. and Fischer, R. F. H and Huber, J.B.},
  journal={IEEE Trans. Commun.},
  volume={52},
  number={12},
  pages={2057-2060},
  year={2004},
  publisher={IEEE}
}

@book{Wan25,
  title={Success Probabilities of the Klein-Babai Estimators and Their Optimization},
  author={Wang, S.},
  year={2025},
  publisher={Master's thesis, McGill University}
}

@inproceedings{sohl2015deep,
  title={Deep unsupervised learning using nonequilibrium thermodynamics},
  author={Sohl-Dickstein, Jascha and Weiss, Eric and Maheswaranathan, Niru and Ganguli, Surya},
  booktitle={International conference on machine learning},
  pages={2256--2265},
  year={2015},
  organization={pmlr}
}

@article{austin2021structured,
  title={Structured denoising diffusion models in discrete state-spaces},
  author={Austin, Jacob and Johnson, Daniel D and Ho, Jonathan and Tarlow, Daniel and Van Den Berg, Rianne},
  journal={Advances in neural information processing systems},
  volume={34},
  pages={17981--17993},
  year={2021}
}

@article{hoogeboom2021argmax,
  title={Argmax flows and multinomial diffusion: Learning categorical distributions},
  author={Hoogeboom, Emiel and Nielsen, Didrik and Jaini, Priyank and Forr{\'e}, Patrick and Welling, Max},
  journal={Advances in neural information processing systems},
  volume={34},
  pages={12454--12465},
  year={2021}
}

@article{zilberstein2022annealed,
  title={Annealed Langevin dynamics for massive {MIMO} detection},
  author={Zilberstein, Nicolas and Dick, Chris and Doost-Mohammady, Rahman and Sabharwal, Ashutosh and Segarra, Santiago},
  journal={IEEE Transactions on Wireless Communications},
  volume={22},
  number={6},
  pages={3762--3776},
  year={2022},
  publisher={IEEE}
}

@inproceedings{he2024massive,
  title={A Massive {MIMO} Sampling Detection Strategy Based on Denoising Diffusion Model},
  author={He, Lanxin and Wang, Zheng and Huang, Yongming},
  booktitle={2024 International Wireless Communications and Mobile Computing (IWCMC)},
  pages={1443--1448},
  year={2024},
  organization={IEEE}
}

@book{BaiCY14,
    author= "L. Bai and J. Choi and Q. Yu",
    year = "2014",
    title = "Low Complexity {MIMO} Receivers",
    publisher = "Springer",
    address = "Cham, Germany",
}

@inproceedings{chang2023success,
  title={Success Probabilities of {L}2-norm Regularized Babai Detectors and Maximization},
  author={Chang, Xiao-Wen and Lu, Qincheng and Xu, Yingzi},
  booktitle={2023 IEEE International Symposium on Information Theory (ISIT)},
  pages={1219--1224},
  year={2023},
  organization={IEEE}
}

@article{ho2020denoising,
  title={Denoising diffusion probabilistic models},
  author={Ho, Jonathan and Jain, Ajay and Abbeel, Pieter},
  journal={Advances in neural information processing systems},
  volume={33},
  pages={6840--6851},
  year={2020}
}

@article{scotti2020graph,
  title={Graph neural networks for massive {MIMO} detection},
  author={Scotti, Andrea and Moghadam, Nima N and Liu, Dong and Gafvert, Karl and Huang, Jinliang},
  journal={arXiv preprint arXiv:2007.05703},
  year={2020}
}

@article{vaswani2017attention,
  title={Attention is all you need},
  author={Vaswani, Ashish and Shazeer, Noam and Parmar, Niki and Uszkoreit, Jakob and Jones, Llion and Gomez, Aidan N and Kaiser, Lukasz and Polosukhin, Illia},
  journal={Advances in neural information processing systems},
  volume={30},
  year={2017}
}

@article{song2020score,
  title={Score-based generative modeling through stochastic differential equations},
  author={Song, Yang and Sohl-Dickstein, Jascha and Kingma, Diederik P and Kumar, Abhishek and Ermon, Stefano and Poole, Ben},
  journal={arXiv preprint arXiv:2011.13456},
  year={2020}
}

@article{sun2023difusco,
  title={Difusco: Graph-based diffusion solvers for combinatorial optimization},
  author={Sun, Zhiqing and Yang, Yiming},
  journal={Advances in neural information processing systems},
  volume={36},
  pages={3706--3731},
  year={2023}
}

@inproceedings{Kle00,
  title={Finding the closest lattice vector when it's unusually close },
  author={Klein, P.},
  booktitle={Proc. ACM-SIAM Symposium on Discrete Algorithms},
  pages={937--941},
  year={2000}
}

@inproceedings{he2016deep,
  title={Deep residual learning for image recognition},
  author={He, Kaiming and Zhang, Xiangyu and Ren, Shaoqing and Sun, Jian},
  booktitle={Proceedings of the IEEE conference on computer vision and pattern recognition},
  pages={770--778},
  year={2016}
}

@article{lou2023discrete,
  title={Discrete diffusion modeling by estimating the ratios of the data distribution},
  author={Lou, Aaron and Meng, Chenlin and Ermon, Stefano},
  journal={arXiv preprint arXiv:2310.16834},
  year={2023}
}

@article{luong2025diffusion,
  title={Diffusion models for future networks and communications: A comprehensive survey},
  author={Luong, Nguyen Cong and Hai, Nguyen Duc and Van Le, Duc and Nguyen, Huy T and Vu, Thai-Hoc and Huynh-The, Thien and Zhang, Ruichen and Anh, Nguyen Duc Duy and Niyato, Dusit and Di Renzo, Marco and others},
  journal={arXiv preprint arXiv:2508.01586},
  year={2025}
}

@inproceedings{barbero2006performance,
  title={Performance analysis of a fixed-complexity sphere decoder in high-dimensional {MIMO} systems},
  author={Barbero, Luis G and Thompson, John S},
  booktitle={2006 IEEE International Conference on Acoustics Speech and Signal Processing Proceedings},
  volume={4},
  pages={IV--IV},
  year={2006},
  organization={IEEE}
}

@article{chang2024extended,
  title={An extended Babai method for estimating linear model based integer parameters},
  author={Chang, Xiao-Wen and Chen, Zhilong and Wen, Jinming},
  journal={Econometrics and Statistics},
  volume={29},
  pages={238--251},
  year={2024},
  publisher={Elsevier}
}

@article{kawar2021snips,
  title={Snips: Solving noisy inverse problems stochastically},
  author={Kawar, Bahjat and Vaksman, Gregory and Elad, Michael},
  journal={Advances in neural information processing systems},
  volume={34},
  pages={21757--21769},
  year={2021}
}

@inproceedings{
song2021denoising,
title={Denoising Diffusion Implicit Models},
author={Jiaming Song and Chenlin Meng and Stefano Ermon},
booktitle={International Conference on Learning Representations},
year={2021},
}

\end{document}